\documentclass[letterpaper, 10 pt, conference]{tex/ieeeconf}  

\IEEEoverridecommandlockouts                              

\usepackage{amsmath}
\usepackage{cite}
\usepackage{amssymb}
\usepackage{multirow}
\usepackage{booktabs}
\usepackage{tikz}
\usepackage{array}
\usepackage{pifont}  
\usepackage{colortbl}  
\usepackage{xcolor}  
\usepackage{flushend}
\usepackage{float}
\usepackage{placeins}
 \usepackage{hyperref}

\usepackage{caption}
\hypersetup{pdfauthor={},pdftitle={Function beyond Form: Functional Correspondence for Cross-Embodiment Dexterous Grasp Generation}}

\definecolor{RankFirstFill}{HTML}{FF9999}
\definecolor{RankSecondFill}{HTML}{FEF8AD}
\newcommand{\RankCell}[2]{\tikz[baseline=(rankvalue.base)]{\node[fill=#1,draw=none,rounded corners=1.2pt,inner xsep=2.4pt,inner ysep=0.7pt,text height=1.55ex,text depth=0.25ex] (rankvalue) {#2};}}
\newcommand{\RankFirst}[1]{\RankCell{RankFirstFill}{\textbf{#1}}}
\newcommand{\RankSecond}[1]{\RankCell{RankSecondFill}{\underline{#1}}}
\newcommand{\RankThird}[1]{#1}
\newcommand{\cmark}{\textcolor{green}{\ding{51}}}  
\newcommand{\xmark}{\textcolor{red}{\ding{55}}}    

\newcommand{\method}{\textsc{FunCo-Grasp}}
\newcommand{\drograsp}{\ensuremath{\mathcal{D}(\mathcal{R},\mathcal{O})} Grasp}
\newcommand{\trograsp}{\ensuremath{\mathcal{T}(\mathcal{R},\mathcal{O})} Grasp}
\title{Function beyond Form: Functional Correspondence for Cross-Embodiment Dexterous Grasp Generation}

\author{
Bolin Zou\textsuperscript{1},
Wenlong Dong\textsuperscript{1},
Mu Ai\textsuperscript{1},
Chao Tang\textsuperscript{2},
Aoxiang Gu\textsuperscript{1},
Lipeng Chen\textsuperscript{3,4},
and Hong Zhang\textsuperscript{1}
\thanks{\textsuperscript{1}Shenzhen Key Laboratory of Robotics
and Computer Vision, Southern University of Science and Technology,
Shenzhen, China.}
\thanks{\textsuperscript{2}Department of Robotics, Perception
and Learning, KTH Royal Institute of Technology,
Stockholm, Sweden.}
\thanks{\textsuperscript{3}School of Artificial Intelligence, Shanghai Jiao Tong University, Shanghai, China.}
\thanks{\textsuperscript{4}Rysen Robotics, Shenzhen, China.}
}

\begin{document}

\maketitle
\thispagestyle{empty}
\pagestyle{empty}

\begin{abstract}
    Cross-embodiment dexterous grasp generation remains challenging because robotic hands differ substantially in geometry, topology, and kinematics. Existing approaches often lack explicit correspondences between structurally different hand regions that play similar functional roles in a grasp, a concept we refer to as functional correspondence. Consequently, their models tend to learn hand-specific interaction patterns rather than transferable grasp knowledge, limiting generalization to unseen hands. To address this limitation, we introduce \method{}, which establishes functional correspondences across heterogeneous hand embodiments. Specifically, Functional Part Alignment aligns each hand to a canonical functional schema by mapping physical links to shared functional parts according to their grasping roles, while Canonical Frame Alignment expresses these parts in canonical local frames. These two alignments provide a consistent representation for inter-part and hand-object interactions, allowing the model to learn transferable grasp knowledge across hands. Conditioned on the aligned hand representation and object geometry, a diffusion model generates the target spatial arrangement of the functional parts, which are then converted into an executable joint configuration. Adapting \method{} to an unseen hand requires only its geometric and kinematic models and a one-time lightweight functional annotation, without target-hand grasp data, fine-tuning, or learned retargeting. In simulation on held-out objects from the filtered CMapDataset, \method{} achieves average success rates of $92.40\%$ on three seen hands and $74.02\%$ on four unseen hands. In real-world experiments, the same model achieves an overall success rate of $76.00\%$ on two unseen hands without additional training or fine-tuning. These results demonstrate the effectiveness of \method{} in transferring grasp knowledge to unseen hands. Project webpage can be found \href{https://zblzz.github.io/FunCo-Grasp-Web/}{here}

\end{abstract}

\section{Introduction}
Dexterous robotic hands exhibit substantial variation in morphology and kinematics, such as finger count, link geometry, and joint topology. Although this diversity supports different manipulation requirements, it also makes grasp knowledge strongly hand-specific and difficult to transfer to unseen hands~\cite{wang2023dexgraspnet,xu2023unidexgrasp}. Consequently, cross-embodiment dexterous grasp generation remains challenging. An ideal system should instead transfer grasp knowledge learned from existing hands to unseen ones with minimal effort. This raises a central question: how can grasp knowledge be structured to capture cross-embodiment correspondences and support transfer to hands with unseen morphologies and kinematics?

\begin{figure}[t]
  \centering
  \vspace*{0.08in}
  \begin{tikzpicture}[inner sep = 0pt, outer sep = 0pt]
    \node[anchor=south west] (fnC) at (0in,0in)
      {\includegraphics[width=\linewidth,clip=true]{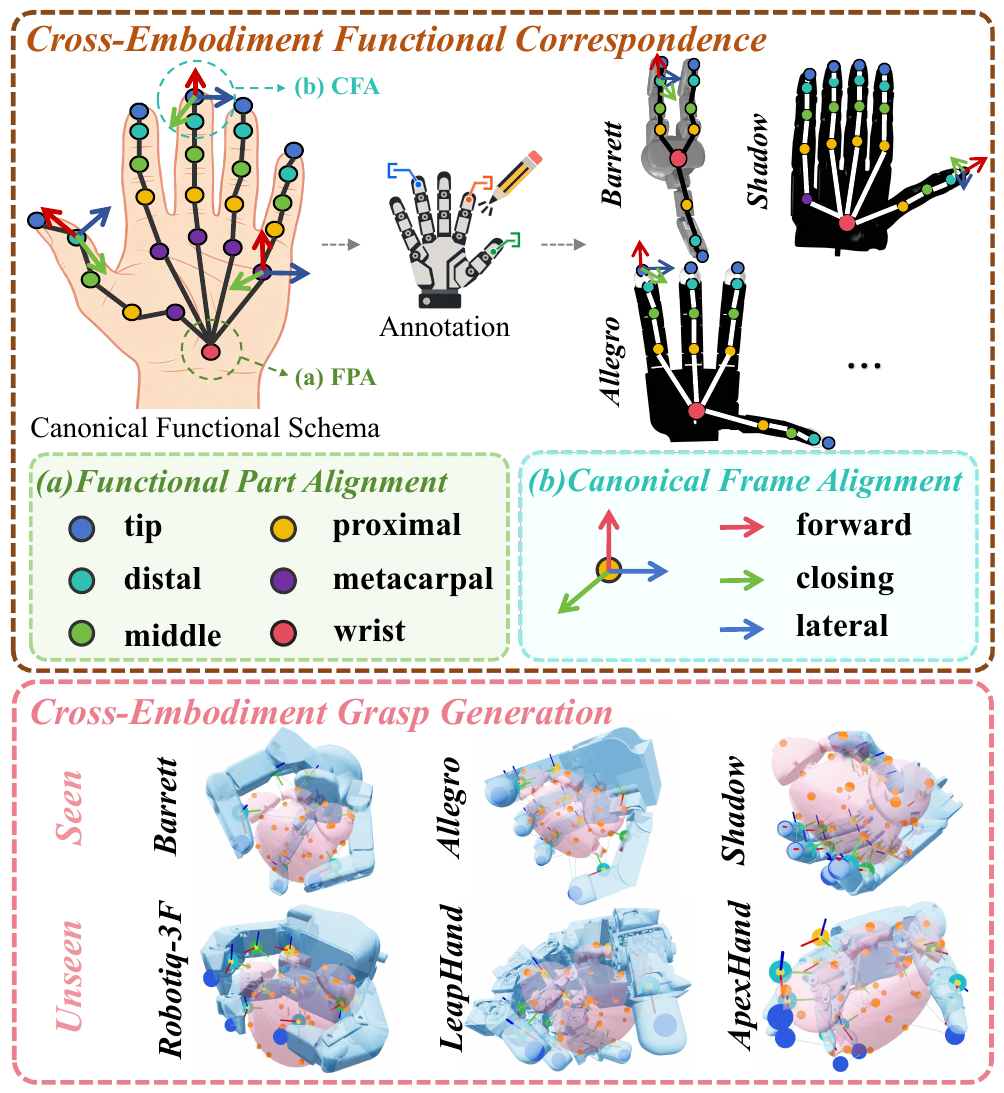}};
  \end{tikzpicture}
  \vspace*{-0.2in}
\caption{\textbf{Overview of \method{}.} Top: The canonical functional schema and cross-embodiment functional correspondences established through a one-time lightweight functional annotation via (a)~Functional Part Alignment and (b)~Canonical Frame Alignment. Bottom: Grasp transfer from seen to unseen hands.}
  \label{fig:intro}
  \vspace*{-0.2in}
\end{figure}

Existing approaches to cross-embodiment dexterous grasp generation can be broadly grouped according to how they represent grasp knowledge: hand-centric, object-centric, and interaction-centric. Hand-centric methods directly encode hand poses or joint configurations within hand-specific state spaces~\cite{wang2023dexgraspnet,xu2024dgtr,lee2026graspgraphnet}, whereas object-centric methods predict contact targets on the object surface~\cite{shao2020unigrasp,attarian2023geomatch,li2023gendexgrasp,wu2025cedex}. Both paradigms focus primarily on one side of a grasp, leaving its coupled hand-object nature implicit. Interaction-centric methods instead represent grasps through explicit hand-object spatial relations~\cite{wei2025dro,fei2025tro,zhang2026mango}. However, these relations are typically defined using hand-specific points, links, or local coordinate frames without explicit correspondences between structurally different hand regions that play similar functional roles in a grasp, a concept we refer to as functional correspondence. Consequently, these methods learn hand-specific interaction patterns rather than transferable grasp knowledge, limiting generalization to unseen hands.

To address this problem, we introduce \method{}, a framework that establishes functional correspondences across heterogeneous hands for cross-embodiment dexterous grasp generation. As shown in Fig.~\ref{fig:intro}, our framework operates across three sequential stages:
(1)~\textbf{Cross-Embodiment Functional Correspondence}, which aligns each hand to a canonical functional schema through two alignments. \textit{Functional Part Alignment} (Fig.~\ref{fig:intro}(a)) maps physical links to shared functional parts according to their grasping roles, while \textit{Canonical Frame Alignment} (Fig.~\ref{fig:intro}(b)) expresses these parts in canonical local frames. These two alignments provide a consistent representation for inter-part and hand-object interactions;
(2)~\textbf{Hand and Object Encoding}, which integrates these functional parts with hand-specific geometry and kinematics into a unified functional node representation on a graph, while extracting geometric features from the object; and
(3)~\textbf{Functional-Part-Based Grasp Generation}, which employs a diffusion model conditioned on the aligned hand representation and object geometry to generate the target spatial arrangement of the functional parts in the shared interaction space before converting them into executable joint configurations via inverse kinematics.
Adapting \method{} to an unseen hand requires only its geometric and kinematic models and a one-time lightweight functional annotation, without target-hand grasp data, fine-tuning, or learned retargeting.


Trained on grasps across 48 objects from three hands in the filtered CMapDataset, \method{} achieves average grasp success rates of $92.40\%$ on three seen hands and $74.02\%$ on four unseen hands in simulation on held-out objects. Furthermore, In real-world experiments, the same model achieves an overall success rate of $76.00\%$ on two unseen hands without additional training or fine-tuning. These results demonstrate the effectiveness of \method{} in transferring grasp knowledge to unseen hands. 

The main contributions of this work are as follows:
\begin{itemize}
    \item We propose a functional-part-based hand representation that establishes functional correspondences for inter-part and hand-object interactions across hands.
    

    \item We propose \method{}, a framework that establishes functional correspondences across heterogeneous hands for cross-embodiment dexterous grasp generation.
    

\end{itemize}

\section{Related Work}

\subsection{Dexterous Grasp Representation}


Existing grasp representation methods fall into three paradigms: hand-centric, object-centric, and interaction-centric. Hand-centric approaches directly represent grasps as wrist poses, joint configurations, or physical link poses in hand-specific state spaces~\cite{wang2023dexgraspnet,xu2023unidexgrasp,xu2024dgtr}. Although these outputs are directly executable, they remain inherently coupled to hand-specific structures, making cross-embodiment transfer challenging. Object-centric approaches instead predict contact points, heatmaps, or correspondences on the object surface to improve cross-embodiment transfer~\cite{li2023gendexgrasp,attarian2023geomatch,wu2025cedex}. However, an object-centric representation primarily specifies where contact should occur on objects, without explicitly modeling the coordinated hand configuration required to realize a stable grasp. Consequently, matching the predicted contacts does not guarantee kinematic feasibility, collision-free finger placement, or favorable grasp geometry.

Interaction-centric representation combines the advantages of both paradigms by explicitly modeling hand-object spatial relations, jointly preserving hand configuration and object contact pattern~\cite{wei2025dro,fei2025tro,zhang2026mango,lee2026graspgraphnet}. However, existing methods typically define spatial relations over hand-specific surface samples, arbitrary links, or local frames, whose correspondences can become ambiguous or even undefined across structurally different hands or hand regions. Consequently, their models learn hand-specific interaction patterns rather than transferable
grasp knowledge, hindering the generalization to unseen hands. In contrast, \method{} establishes functional correspondences across heterogeneous hands, enabling grasp knowledge transfer to unseen hands without target-hand grasp data, fine-tuning, or learned retargeting.

\begin{figure*}[t]
  \centering
  \includegraphics[height=3.5in,clip=true]{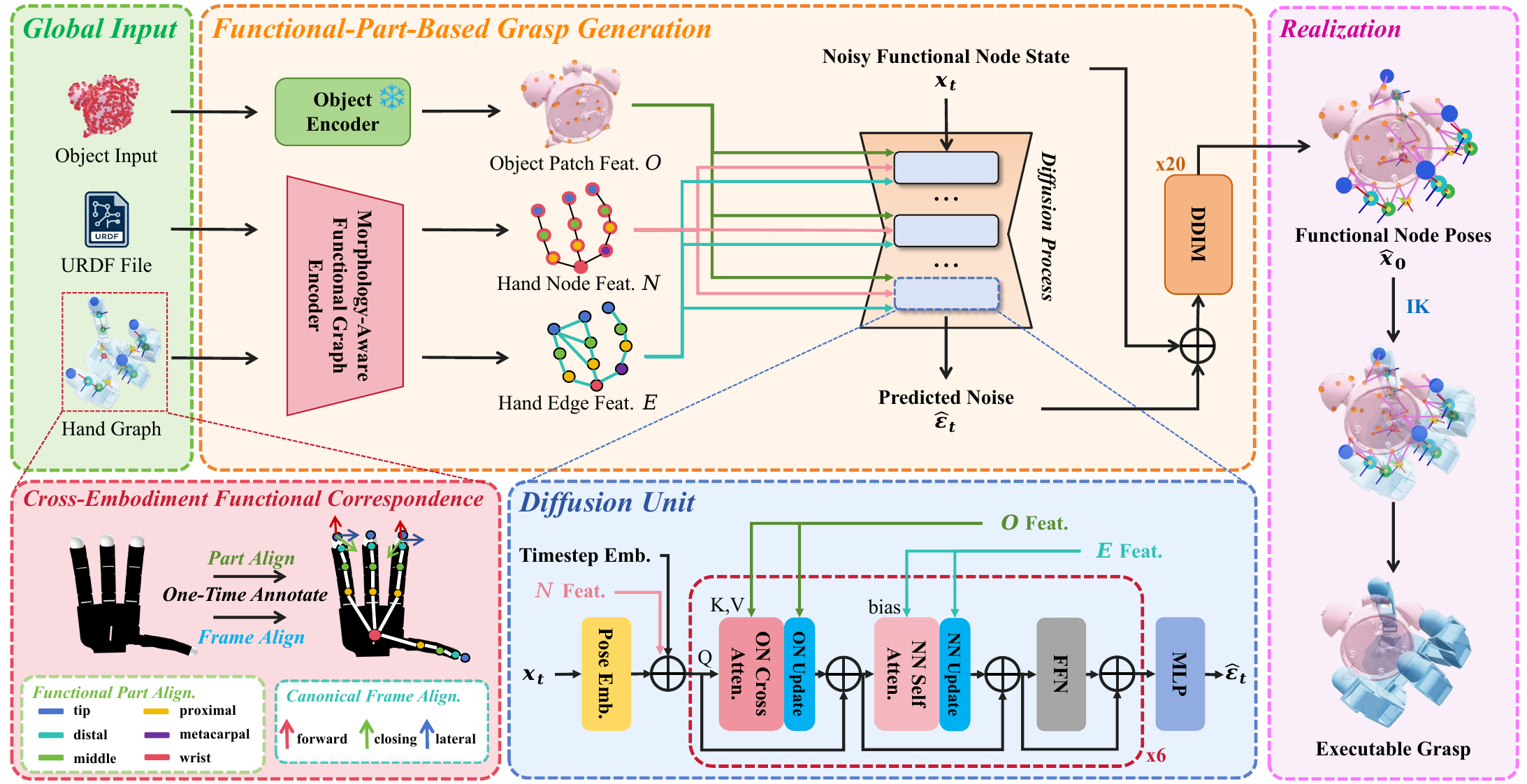}
  \caption{\textbf{Overview of \method{}.} (1)~\textbf{Cross-Embodiment Functional Correspondence} maps physical links to shared functional parts according to their grasping roles and expresses them in canonical local frames. (2)~\textbf{Hand and Object Encoding} extracts node features \(\mathbf{N}\), edge features \(\mathbf{E}\), and object patch features \(\mathbf{O}\). (3)~\textbf{Functional-Part-Based Grasp Generation} conditions on the aligned hand representation and object geometry to generate the target spatial arrangement of the functional parts in the shared interaction space, before converting them into executable joint configurations via inverse kinematics.}
  \label{fig:overview}
  \vspace*{-0.2in}
\end{figure*}

\subsection{Morphology-Aware Cross-Embodiment Learning}

Beyond grasp representation, cross-embodiment transfer also depends on how the model uses hand structures. For manipulation and control, existing methods adapt to heterogeneous hands through shared action spaces~\cite{yuan2025crossdex} or kinematic graph neural networks~\cite{patel2024getzero}. For grasp generation, prior approaches map distinct hands to a canonical kinematic pose space~\cite{wu2026unimorph} or condition generation on hand morphology to predict low-dimensional articulation coefficients~\cite{zhang2025machagrasp}. These methods use hand structures primarily to organize joint articulations, control commands, or canonical pose spaces.

Closest to our work are DexGrasp-Zero~\cite{wu2026dexgraspzero} and UniMorphGrasp~\cite{wu2026unimorph}, which use morphology alignment for cross-embodiment transfer. The former aligns hand structures to unify control and joint-motion spaces, while the latter maps active joints to canonical slots and masks missing joints. These methods primarily establish correspondences for joint motion or joint configurations. In contrast, \method{} establishes functional correspondences for inter-part and hand-object interactions by mapping physical links of hand regions to shared functional parts according to their grasping roles. A functional part can include multiple physical links and is expressed in a canonical local frame. This alignment provides a consistent representation for these interactions, allowing the model to transfer grasp knowledge learned from existing hands to unseen hands without target-hand grasp data, fine-tuning, or learned retargeting.



\section{Method}

To learn transferable grasp knowledge via functional correspondences, \method{} operates across three sequential stages (Fig.~\ref{fig:overview}): (1)~Cross-Embodiment Functional Correspondence, (2)~Hand and Object Encoding, and (3)~Functional-Part-Based Grasp Generation.

\subsection{Problem Formulation}
We consider cross-embodiment dexterous grasp generation for a target hand that may be unseen during training. Given an object point cloud \(\mathbf{P}\subset\mathbb R^3\), the target hand's geometric and kinematic models \(\mathbf{H}\) provided by its URDF and associated geometry, and a one-time lightweight functional annotation \(\mathbf{A}\), the goal is an executable grasp:
\begin{equation}
    (\mathbf{P},\mathbf{H},\mathbf{A})\longmapsto(\mathbf{T}_w,\mathbf{q})\in\mathrm{SE}(3)\times\mathbb R^D.
\label{eq:problem}
\end{equation}
Here, \(\mathbf{T}_w\) is the wrist pose in the object frame, and \(\mathbf{q}\) contains the target hand's \(D\) internal joint angles, excluding the wrist degrees of freedom. The grasp should stably hold the object while satisfying the hand's kinematic constraints and joint limits. We seek to transfer grasp knowledge learned from existing hands to unseen hands without target-hand grasp data, fine-tuning, or learned retargeting.

\begin{figure}[t]
  \centering
  \includegraphics[width=\linewidth,clip=true]{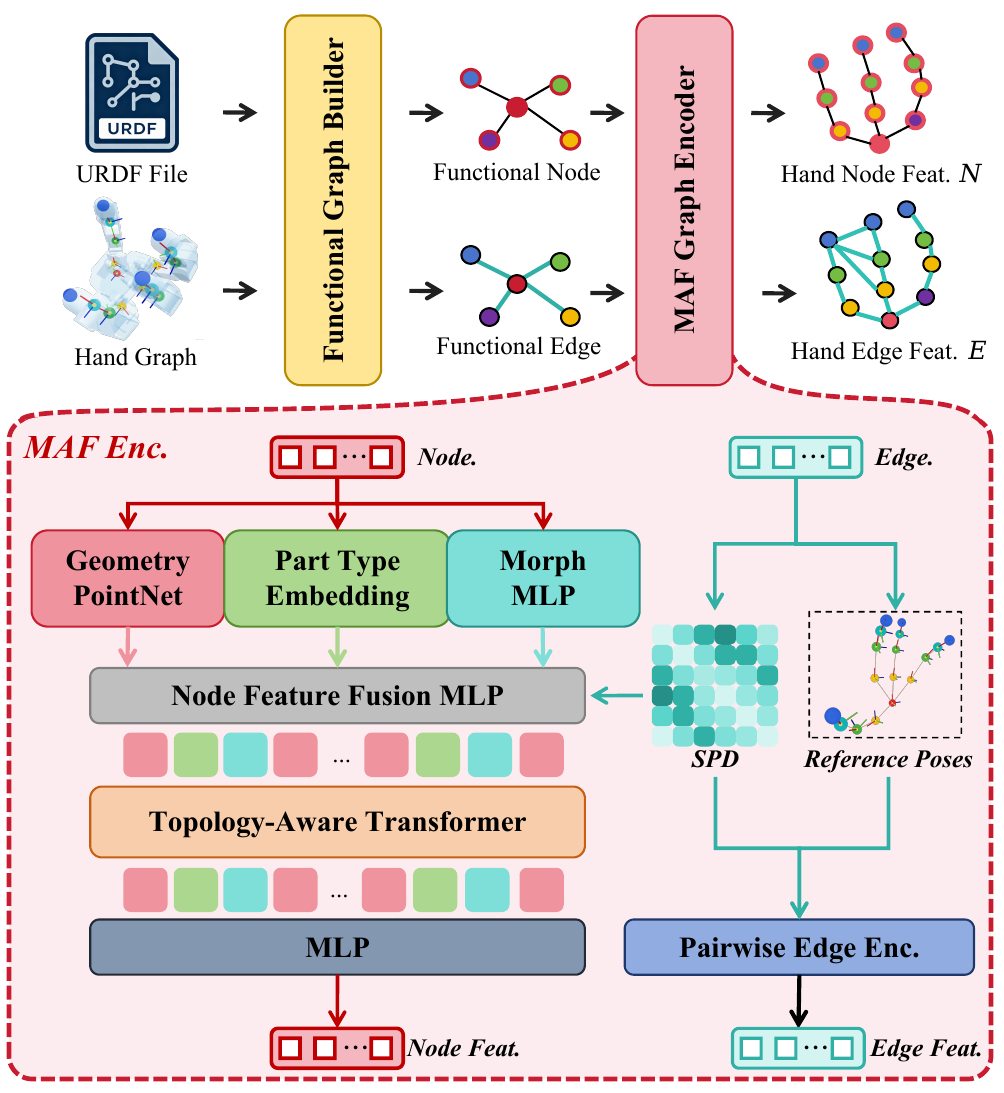}
  \caption{\textbf{MAF Graph Encoder.} Given the URDF and hand graph, the node branch encodes part geometry, grasping roles, and kinematics into \(\mathbf{N}\). The edge branch encodes reference part poses and kinematic connections into \(\mathbf{E}\).}
  \label{fig:maf}
  \vspace*{-0.2in}
\end{figure}

\subsection{Cross-Embodiment Functional Correspondence}
To establish functional correspondences across heterogeneous hands, we construct a functional-part-based hand representation. Specifically, as shown in Fig.~\ref{fig:intro} \textit{Functional Part Alignment} maps physical links of hand regions to shared functional parts according to their grasping roles, while \textit{Canonical Frame Alignment} expresses these parts in canonical local frames. Together, these alignments provide a consistent representation for inter-part and hand-object interactions across hands.

\textbf{Functional Part Alignment.} We define functional parts as link groups assigned to the canonical functional schema (Fig.~\ref{fig:intro}) according to their grasping roles and equipped with canonical local frames. The schema is illustrated in the upper part of Fig.~\ref{fig:intro}, with shared wrist, metacarpal, proximal, middle, distal, and tip types in Fig.~\ref{fig:intro}(a); digit labels further distinguish parts on different fingers. The one-time lightweight functional annotation \(\mathbf{A}\) assigns physical links to these parts and marks their surface regions for interaction. Each part thus comprises its associated links, surface point cloud, and URDF-derived geometric and kinematic attributes. A part may contain multiple links, and absent parts may be omitted. This mapping therefore establishes explicit correspondences between structurally different hand regions that play similar functional roles in a grasp.

\textbf{Canonical Frame Alignment.} After identifying corresponding functional parts, we align their local coordinate frames. Each canonical local frame uses forward, closing, and lateral axes (Fig.~\ref{fig:intro}(b)). The annotated interaction region's weighted centroid defines the origin. The forward and closing axes follow the part's extension and closing motion, respectively. We orthonormalize these two axes in the link frame that defines the part's orientation and use their cross product as the lateral axis. Forward kinematics updates this frame during motion. The wrist pose uses the wrist part's canonical frame, with wrist-relative poses from left and right hands converted to the same left-hand convention. Overall, two alignments provide a consistent representation for inter-part and hand-object interactions, allowing the model to learn transferable grasp knowledge across hands.


\subsection{Hand and Object Encoding}

We encode the hand and object in separate branches to condition grasp generation on the aligned hand representation and object geometry. In the graph diffusion model, each functional part is represented as a functional node, and adjacent parts are connected along the target hand's kinematic tree, forming the hand graph shown in Fig.~\ref{fig:overview}. We sample a valid reference configuration independently of the object and target grasp, then use forward kinematics with \(\mathbf{H}\) and \(\mathbf{A}\) to compute reference part poses \(\mathbf{R}\), including positions and orientations in the common wrist-relative convention. The Morphology-Aware Functional (MAF) Graph Encoder in Fig.~\ref{fig:maf} produces node features \(\mathbf{N}\) and edge features \(\mathbf{E}\), while the object branch produces patch features \(\mathbf{O}\).

\textbf{Node Features.} The node encoder \(f_{\mathrm N}\) combines each functional part's grasping role, geometry, and kinematics. PointNet encodes the part point cloud and annotated interaction region~\cite{qi2017pointnet}. These geometric features are fused with URDF attributes and part labels, then aggregated across the hand graph into a unified functional node representation:
\begin{equation}
    \mathbf{N}=f_{\mathrm N}(\mathbf{H},\mathbf{A}).
\label{eq:node_encoding}
\end{equation}

\textbf{Edge Features.} To encode inter-part relations, the edge encoder \(f_{\mathrm E}\) combines relative positions and orientations computed from \(\mathbf{R}\) with shortest-path distances on hand graph:
\begin{equation}
    \mathbf{E}=f_{\mathrm E}(\mathbf{H},\mathbf{A},\mathbf{R}).
\label{eq:edge_encoding}
\end{equation}
Together, \(\mathbf{N}\) and \(\mathbf{E}\) encode the aligned hand representation while retaining hand-specific geometry and kinematics.

\textbf{Object Features.} Following TRO-Grasp~\cite{fei2025tro}, a frozen VQ-VAE extracts local patch descriptors from \(\mathbf{P}\). Combining these descriptors with patch centers and object scale forms \(\mathbf{O}\). The conditions \(\mathbf{N}\), \(\mathbf{E}\), \(\mathbf{O}\), and \(\mathbf{R}\) remain fixed during each grasp-sampling run.


\subsection{Functional-Part-Based Grasp Generation}
Conditioned on the aligned hand representation and object geometry, a diffusion model generates the target spatial arrangement of the functional parts in the shared interaction space. Each functional node uses translation and axis-angle rotation: the wrist pose is expressed in the object frame, while other nodes encode residuals from \(\mathbf{R}\) in the common wrist-relative convention. Translations are normalized by object scale, with wrist translation also centered on the object; invalid and padded channels are masked. The noisy state \(\mathbf{x}_t\) at timestep \(t\) specifies the current part arrangement.

\textbf{Interaction Modeling.} At each denoising step, we compose the residuals in \(\mathbf{x}_t\) with \(\mathbf{R}\) and apply the current wrist pose to recover part poses in the object frame. Object--Node (ON) relations express object patch centers in each part's canonical local frame, together with part orientation and validity; Node--Node (NN) relations describe relative part translations, rotations, and validity. 

The denoiser \(f\) models hand-object interactions through ON cross-attention and inter-part interactions through NN self-attention. Conditioned on \(\mathbf{N}\), \(\mathbf{E}\), and \(\mathbf{O}\), it predicts noise from these relations, the part arrangement, and the timestep:
\begin{equation}
    \hat{\epsilon}=f(\mathbf{x}_t,t;\mathbf{N},\mathbf{E},\mathbf{O},\mathbf{R}).
\label{eq:noise_prediction}
\end{equation}
Using the predicted noise, DDIM~\cite{song2021ddim} updates the arrangement from timestep \(t\) to the next sampled timestep \(t'<t\):
\begin{equation}
    \mathbf{x}_{t'}=\operatorname{DDIM}(\mathbf{x}_t,\hat{\epsilon};t,t').
\label{eq:generation}
\end{equation}
Starting from standard Gaussian noise on valid channels, repeated stochastic DDIM updates produce \(\mathbf{x}_0\). 

\begin{table*}[t]
\centering
\begingroup
\setlength{\tabcolsep}{4.0pt}
\renewcommand{\arraystretch}{1.14}
\resizebox{\textwidth}{!}{%
\begin{tabular}{@{}l|ccc|c|cccc|c|c@{}}
\toprule
\multirow{3}{*}{Method}
& \multicolumn{9}{c|}{Success Rate (\%) $\uparrow$}
& \multirow{3}{*}{Eff. (s) $\downarrow$} \\
\cmidrule{2-10}
& \multicolumn{4}{c|}{Seen Hands}
& \multicolumn{5}{c|}{Unseen Hands}
& \\
\cmidrule{2-5}\cmidrule{6-10}
& Shadow & Allegro & Barrett & Avg.
& ApexHand & XHand & LeapHand & Robotiq-3F & Avg. & \\
\midrule
GenDexGrasp~\cite{li2023gendexgrasp}
& 53.30 & 67.00 & 50.00 & 56.77
& 15.17 & 29.37 & \RankThird{25.33} & 31.00 & 25.22
& 18.77 \\
CEDex~\cite{wu2025cedex}
& 86.43 & \RankSecond{92.66} & 85.16 & 88.08
& \RankSecond{46.20} & \RankSecond{62.00} & \RankSecond{71.50} & \RankFirst{86.30} & \RankSecond{66.50}
& 12.52 \\
$\mathcal{D(R,O)}$ Grasp~\cite{wei2025dro}
& 77.80 & 89.90 & \RankThird{88.30} & 85.33
& \RankThird{24.20} & \RankThird{58.60} & 4.80 & \RankThird{64.90} & \RankThird{38.13}
& \RankThird{0.75} \\
$\mathcal{T(R,O)}$ Grasp~\cite{fei2025tro}
& \RankThird{92.50} & 89.10 & \RankSecond{91.60} & \RankThird{91.07}
& 1.30 & 4.00 & 0.10 & 42.40 & 11.95
& \RankSecond{0.21} \\
UniMorphGrasp$^{\dagger}$~\cite{wu2026unimorph}
& \RankFirst{98.80} & \RankThird{90.30} & \RankFirst{93.00} & \RankFirst{94.00}
& -- & -- & -- & -- & --
& 0.45 \\
\midrule
\method{} (Ours)
& \RankSecond{93.80} & \RankFirst{97.20} & 86.20 & \RankSecond{92.40}
& \RankFirst{68.20} & \RankFirst{72.73} & \RankFirst{80.40} & \RankSecond{74.73} & \RankFirst{74.02}
& \RankFirst{0.16} \\
\bottomrule
\end{tabular}%
}
\endgroup
\caption{Grasp success on held-out objects and per-grasp runtime. Red and yellow mark the best and second-best values. $^{\dagger}$UniMorphGrasp results come from its paper because its code is not publicly available. -- denotes unavailable results.}
\label{tab:overall}
\end{table*}

\begin{figure*}[t]
  \centering
  \includegraphics[height=2.6in,clip=true]{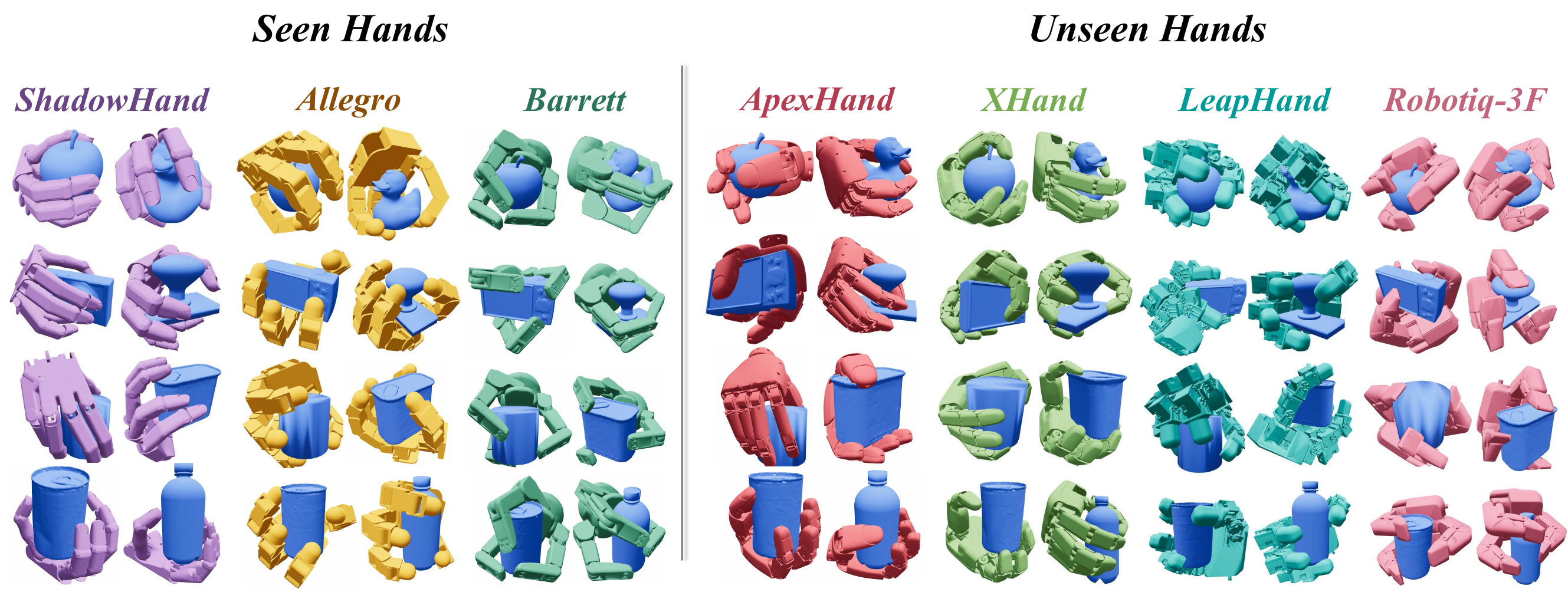}
  \caption{Grasps generated by \method{} on held-out objects with seen (left) and unseen (right) hands. The same model transfers to unseen hands without target-hand grasp data, fine-tuning, or learned retargeting.}
  \label{fig:exe}
  \vspace*{-0.2in}
\end{figure*}

\textbf{Training Objective.} We learn part arrangements through noise prediction, with penetration and contact losses supervising their geometry. The training objective averages over grasps and sampled timesteps:
\begin{equation}
    L = L_d + \mathbb{E}\!\left[w_t(\lambda_p L_p+\lambda_c L_c)\right].
\label{eq:training_objective}
\end{equation}
The noise loss sums four masked mean squared errors between predicted noise \(\hat{\epsilon}\) and injected Gaussian noise \(\epsilon\)~\cite{ho2020ddpm}:
\begin{equation}
    L_d=\sum_{k=1}^{4}\frac{\|m_k\odot(\hat{\epsilon}_k-\epsilon_k)\|_2^2}{\|m_k\|_1}.
\label{eq:noise_loss}
\end{equation}
The four blocks cover wrist and non-wrist translations and rotations; the binary mask \(m_k\) selects valid scalar entries across the batch, and \(\odot\) denotes elementwise multiplication. Separate normalization prevents the more numerous non-wrist channels from diluting wrist supervision

Geometric losses supervise the denoised part arrangement via the object signed distance field \(\Phi_{\mathbf{P}}\)~\cite{turpin2022graspd,zhong2025dexgraspanything}:
\begin{equation}
    L_p = \sum_j a_j \max(-\Phi_{\mathbf{P}}(\mathbf{v}_j), 0), \quad L_c = \sum_{k \in \mathcal{C}} |\Phi_{\mathbf{P}}(\mathbf{v}_k)|.
\label{eq:geo_loss}
\end{equation}
penalizing surface points \(\mathbf{v}_j\) penetrating the object with area weights \(a_j\), and encouraging contacts at points indexed by \(\mathcal{C}\). The timestep weight \(w_t=\bar\alpha_t^2\) weakens geometric supervision under heavy noise.

\textbf{Realization.} Finally, we convert the target spatial arrangement of the functional parts into executable joint configurations via inverse kinematics. We adapt PyRoki~\cite{kim2025pyroki} to our functional node representation, fitting positions of functional parts rather than individual physical links:
\begin{equation}
    (\mathbf{T}_w,\mathbf{q})=\operatorname*{arg\,min}_{\mathbf{T}_w,\mathbf{q}}\sum_i w_i\bigl\|\operatorname{FK}_i(\mathbf{T}_w,\mathbf{q})-\mathbf{p}_i\bigr\|^2.
\label{eq:ik}
\end{equation}
Here, \(\mathbf{p}_i\) is the target part position recovered from \(\mathbf{x}_0\) and \(\mathbf{R}\). Using \(\mathbf{H}\) and \(\mathbf{A}\), \(\operatorname{FK}_i\) computes the weighted centroid of the part's annotated interaction region, and \(w_i\) weights its position error. The sum covers valid functional parts. Canonical local frames provide a consistent representation for inter-part and hand--object interactions during grasp generation, while inverse kinematics fits part positions without explicitly matching their orientations. Adapting \method{} to an unseen hand requires only its geometric and kinematic models and a one-time lightweight functional annotation, without target-hand grasp data, fine-tuning, or learned retargeting.

\section{Experiments}

We evaluate whether functional correspondences enable \method{} to transfer grasp knowledge to unseen hands. Our evaluation includes simulation comparisons, alignment ablations, finger-length variations, diversity analysis, and real-world experiments.

\subsection{Experimental Setup}
\textbf{Dataset and Hands.} We use the filtered CMapDataset split of 48 training and 10 held-out test objects from ContactDB and YCB~\cite{li2023gendexgrasp,wei2025dro,fei2025tro,brahmbhatt2019contactdb,calli2015ycb}. We train on 14,011 grasps with balanced sampling across hands: 3,754 from Allegro, 6,242 from Barrett, and 4,015 from ShadowHand. We evaluate these three seen hands and four unseen hands (ApexHand, XHand, LeapHand, and Robotiq-3F) on the same held-out objects. Adapting the model to each unseen hand requires only its geometric and kinematic models and a one-time lightweight functional annotation.

\begin{figure}[t]
  \centering
  \includegraphics[width=\linewidth,clip=true]{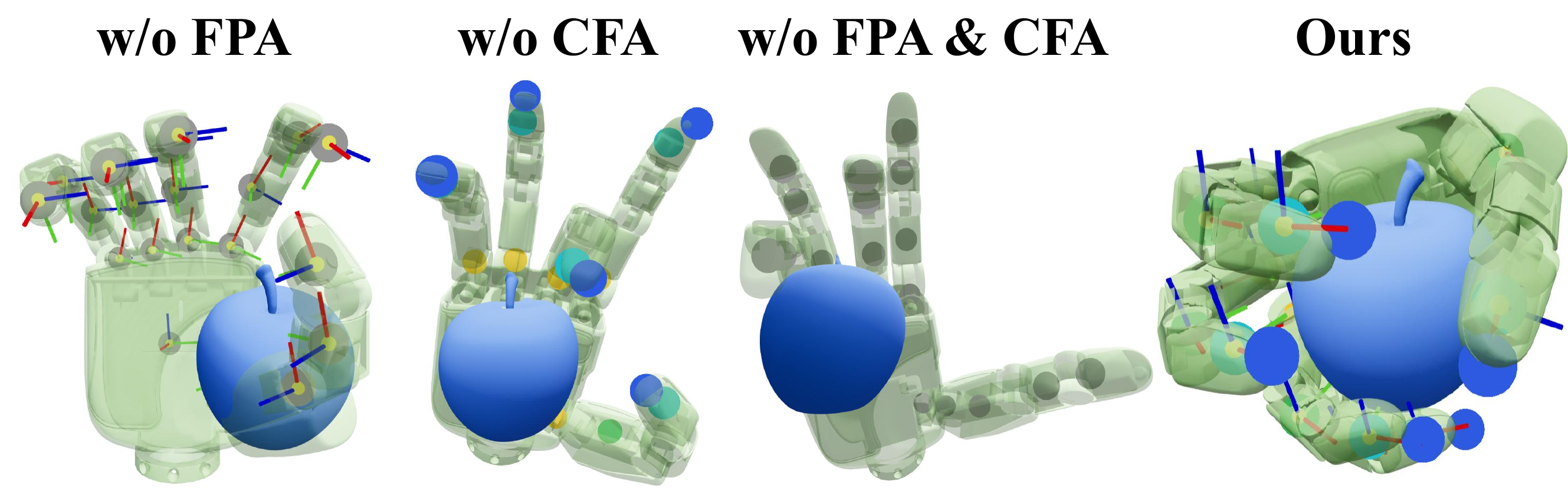}
  \caption{Effect of FPA and CFA on ApexHand grasps: without FPA, hand regions are misplaced; without CFA, coordinated closure is disrupted.}
  \label{fig:alignment_ablation}
  \vspace*{-0.2in}
\end{figure}

\begin{table*}[t]
\centering
\begingroup
\setlength{\tabcolsep}{4.2pt}
\renewcommand{\arraystretch}{1.14}
\resizebox{\textwidth}{!}{%
\begin{tabular}{@{}l|cc|ccc|c|cccc|c@{}}
\toprule
\multirow{3}{*}{Representation}
& \multirow{3}{*}{FPA}
& \multirow{3}{*}{CFA}
& \multicolumn{9}{c}{Success Rate (\%) $\uparrow$} \\
\cmidrule{4-12}
& & & \multicolumn{4}{c|}{Seen Hands}
& \multicolumn{5}{c}{Unseen Hands} \\
\cmidrule{4-7}\cmidrule{8-12}
& & & Shadow & Allegro & Barrett & Avg.
& ApexHand & XHand & LeapHand & Robotiq-3F & Avg. \\
\midrule
Full \method{} & \cmark & \cmark
& 93.80 & \textbf{97.20} & 86.20 & \textbf{92.40}
& \textbf{68.20} & \textbf{72.73} & \textbf{80.40} & \textbf{74.73} & \textbf{74.02} \\
\midrule
w/o FPA & \xmark & \cmark
& \textbf{94.20} & 93.00 & 85.00 & 90.73
& 13.40 & 45.20 & 9.20 & 67.80 & 33.90 \\
w/o CFA & \cmark & \xmark
& 91.60 & 92.00 & 86.80 & 90.13
& 10.40 & 17.00 & 20.20 & 30.00 & 19.40 \\
w/o FPA \& CFA & \xmark & \xmark
& 89.80 & 94.60 & \textbf{87.80} & 90.73
& 3.00 & 0.00 & 0.80 & 0.00 & 0.95 \\
\bottomrule
\end{tabular}%
}
\endgroup
\caption{Effect of Functional Part Alignment (FPA) and Canonical Frame Alignment (CFA) on grasp success across seen and unseen hands. Best results in each column are bold.}
\label{tab:alignment_ablation}
\end{table*}

\textbf{Baselines.} Our baselines cover the grasp representation and morphology-alignment approaches discussed in Related Work. The object-centric baselines are GenDexGrasp~\cite{li2023gendexgrasp} and CEDex~\cite{wu2025cedex}; the interaction-centric baselines are \drograsp{}~\cite{wei2025dro} and \trograsp{}~\cite{fei2025tro}. For these four baselines, we use the released checkpoints without retraining and retain their generation, realization, and control settings. UniMorphGrasp~\cite{wu2026unimorph} represents the morphology-alignment approach. Its code is not publicly available, so we take its seen-hand success rates, runtime, and diversity directly from the paper. These values follow its original evaluation protocol; results for our four unseen hands are unavailable and marked --.

\textbf{Simulation Protocol.} For each hand--object pair, we conduct three evaluation runs, generating 100 new grasps in each run. In Isaac Gym~\cite{makoviychuk2021isaacgym}, the position controller closes the hand for 1~s, followed by six directional disturbances, each lasting 1~s with force magnitude equal to object mass times $0.5\,\mathrm{m\,s^{-2}}$. Success requires final object displacement below 2~cm~\cite{wei2025dro,fei2025tro}.

\textbf{Evaluation Metrics.} Success is averaged equally over repetitions, objects, and hands. Diversity measures the standard deviation of successful configurations, including six floating wrist degrees of freedom and finger joints~\cite{wei2025dro,fei2025tro}. For all locally evaluated methods (including CEDex), per-grasp runtime uses batch size 1 and includes encoding, generation, and optimization or inverse kinematics. Measurements use an RTX 5880 Ada workstation after warm-up and GPU synchronization. Our inverse kinematics uses CPU-based PyRoki adapted to the functional node representation. We exclude loading, one-time compilation, I/O, and simulation.

\textbf{Implementation Details.} The model uses 25 functional node slots, 256-dimensional node features, 64-dimensional ON/NN edge states, and six diffusion units. It has 13.647M parameters: 13.041M trainable and 0.606M in the frozen VQ-VAE. We train for 300 epochs in FP32 on two 48~GB NVIDIA RTX 5880 Ada GPUs, with 24 grasps per GPU. We use Adam with an initial learning rate of $1.41\times10^{-4}$, reduced by a factor of 0.8 every 20 epochs. For each training grasp, we sample ten diffusion timesteps and the corresponding Gaussian noise. All results use the epoch-300 checkpoint, 20-step stochastic DDIM, and the inverse kinematics described in Eq.~\ref{eq:ik}. DDIM uses \(\eta=1\) and scales each step's injected noise by 0.2; initial noise is standard Gaussian.

\begin{figure}[t]
  \centering
  \includegraphics[width=\linewidth,clip=true]{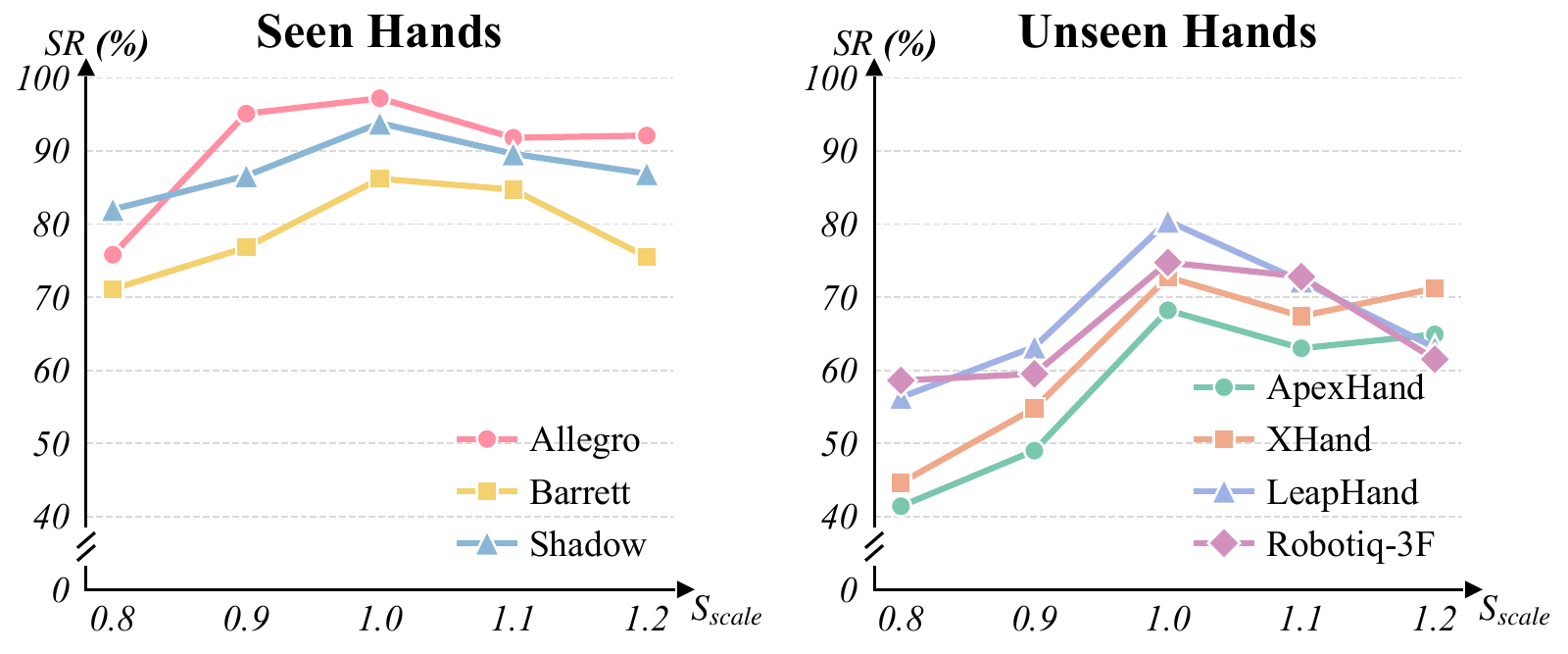}
  \caption{Grasp success under finger-length scaling for seen (left) and unseen (right) hands, without retraining. $1.0\times$ denotes original geometry.}
  \label{fig:morphology_robustness}
  \vspace*{-0.2in}
\end{figure}

\subsection{Simulation Results}
\textbf{Seen-Hand Performance.} On seen hands, \method{} achieves $92.40\%$ average success, compared with $91.07\%$ for \trograsp{} and $85.33\%$ for \drograsp{} under local evaluation (Tab.~\ref{tab:overall}). It reaches $97.20\%$ on Allegro and $93.80\%$ on ShadowHand, exceeding CEDex and \trograsp{} by $4.54$ and $1.30$ percentage points, respectively. These results show that the aligned hand representation maintains high grasp success on seen hands, with grasp examples in Fig.~\ref{fig:exe} (left).

\textbf{Transfer to Unseen Hands.} On unseen hands, \method{} achieves $74.02\%$ average success, compared with $66.50\%$ for CEDex, $38.13\%$ for \drograsp{}, and $11.95\%$ for \trograsp{} (Tab.~\ref{tab:overall}). It leads on three of four unseen hands, exceeding CEDex by $22.00$, $10.73$, and $8.90$ percentage points on ApexHand, XHand, and LeapHand, respectively. These results and the grasp examples in Fig.~\ref{fig:exe} (right) demonstrate transfer of grasp knowledge to unseen hands without target-hand grasp data, fine-tuning, or learned retargeting.

\textbf{Generation Efficiency.} \method{} generates an executable grasp, including inverse kinematics, in $0.16$~s, compared with $0.21$~s for \trograsp{}, $0.75$~s for \drograsp{}, and $12.52$~s for CEDex (Tab.~\ref{tab:overall}). This reduces runtime by $23.81\%$ relative to \trograsp{} and provides a $4.69\times$ speedup over \drograsp{}.

\begin{table}[t]
\centering
\begingroup
\setlength{\tabcolsep}{4.0pt}
\renewcommand{\arraystretch}{1.14}
\resizebox{\columnwidth}{!}{%
\begin{tabular}{@{}l|ccc|c@{}}
\toprule
Method & Shadow & Allegro & Barrett & Avg. \\
\midrule
DFC~\cite{liu2022dfc}
& 0.435 & 0.454 & 0.532 & 0.474 \\
GenDexGrasp~\cite{li2023gendexgrasp}
& 0.318 & 0.389 & 0.488 & 0.398 \\
CEDex~\cite{wu2025cedex}
& 0.438 & \RankFirst{0.473} & \RankSecond{0.624} & \RankSecond{0.512} \\
\drograsp{}~\cite{wei2025dro}
& \RankSecond{0.441} & 0.397 & 0.513 & 0.450 \\
\trograsp{}~\cite{fei2025tro}
& 0.292 & 0.370 & 0.542 & 0.401 \\
UniMorphGrasp$^{\dagger}$~\cite{wu2026unimorph}
& \RankFirst{0.445} & \RankSecond{0.462} & \RankFirst{0.698} & \RankFirst{0.535} \\
\midrule
\method{} (Ours)
& 0.282 & 0.331 & 0.268 & 0.293 \\
\bottomrule
\end{tabular}%
}
\endgroup
\caption{Diversity of successful grasps on seen hands (rad.). Baselines use paper-reported values under their published protocols; $^{\dagger}$UniMorphGrasp uses its full model.}
\label{tab:diversity}
\vspace*{-0.15in}
\end{table}

\begin{figure}[t]
  \centering
  \includegraphics[width=\linewidth,clip=true]{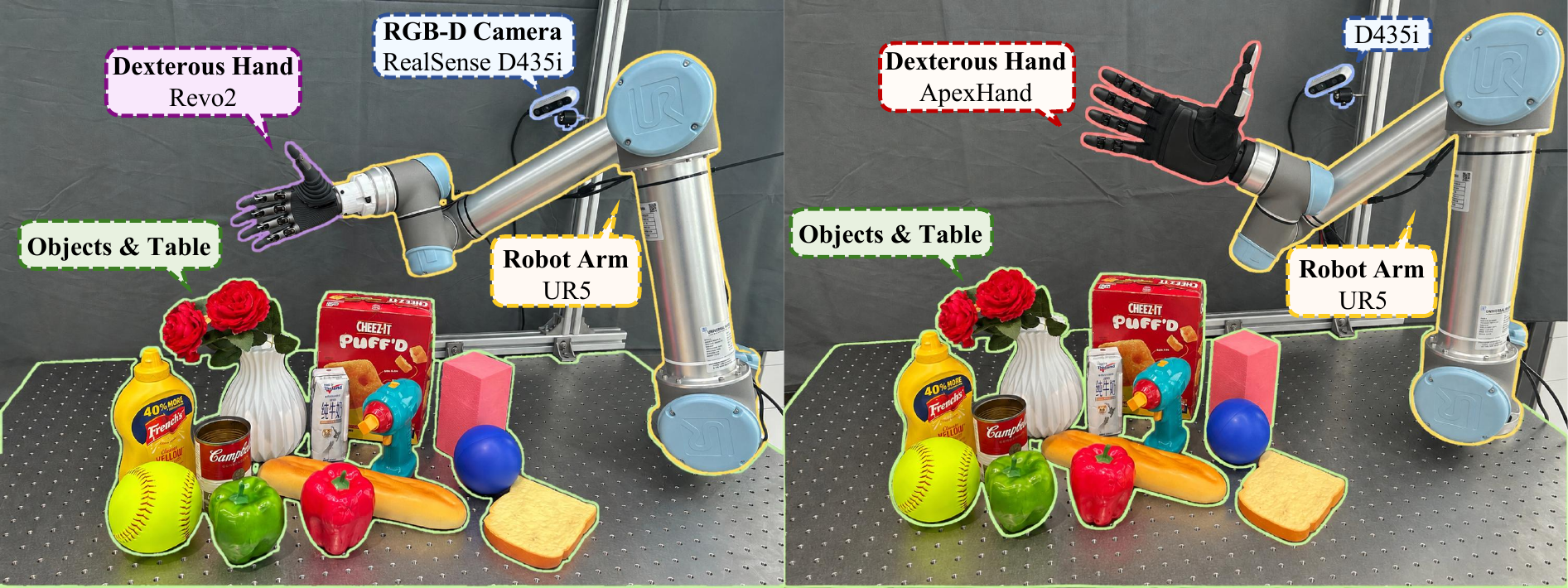}
  \caption{Real-world setups with Revo2 (left) and ApexHand (right) mounted on UR5 arms, using RealSense D435i observations. Both hands are unseen during training.}
  \label{fig:real_setup}
  \vspace*{-0.2in}
\end{figure}

\subsection{Ablation Studies}
\textbf{Alignment Variants.} Tab.~\ref{tab:alignment_ablation} compares variants of \textit{Functional Part Alignment} (FPA) and \textit{Canonical Frame Alignment} (CFA). Without FPA, we replace functional parts with physical-link nodes, using whole-link surfaces, generic part and digit labels, and the physical-link graph while retaining canonical local axes. Without CFA, we retain functional parts and their centers but use the native URDF axes of the links that define part orientation. Local geometry, part poses, and interaction relations follow this frame definition consistently, while the diffusion state parameterization remains unchanged. The final variant combines both changes. We train all variants from scratch with the same training data and optimization settings.

\textbf{Alignment Effects.} Removing FPA or CFA reduces unseen-hand success from $74.02\%$ to $33.90\%$ and $19.40\%$, respectively; removing both reduces it to $0.95\%$ (Tab.~\ref{tab:alignment_ablation}). The individual drops are $40.12$ and $54.62$ percentage points, whereas seen-hand success decreases from $92.40\%$ to $90.13$--$90.73\%$ across the ablated variants. This contrast shows the importance of functional correspondences for transferring grasp knowledge to unseen hands.

\textbf{Complementary Roles.} The ApexHand examples in Fig.~\ref{fig:alignment_ablation} illustrate the complementary roles of the two alignments. Without FPA, fingers tend to close but hand regions are misplaced; without CFA, coordinated closure is disrupted. Success decreases from \(68.20\%\) to \(13.40\%\) and \(10.40\%\), respectively (Tab.~\ref{tab:alignment_ablation}). Together, shared functional parts and canonical local frames provide a consistent representation for inter-part and hand--object interactions.

\subsection{Robustness to Finger-Length Variations}
\textbf{Finger-Length Variation.} We evaluate all seven hands under finger-length scaling in Fig.~\ref{fig:morphology_robustness}. Finger links, fingertips, and downstream joint offsets are scaled along the forward axis by $S_{\mathrm{scale}}\in\{0.8,0.9,1.0,1.1,1.2\}$. Palm and wrist geometry, palm-to-finger-base offsets, topology, joint axes and limits, and canonical-frame orientations remain fixed. Geometry-dependent attributes are recomputed without retraining or redefining functional correspondence.

\textbf{Robustness Results.} At $1.1\times$ finger length, success exceeds $84\%$ on all seen hands and reaches at least $63\%$ on all unseen hands; at $1.2\times$, all seen and unseen hands remain above $75\%$ and $60\%$, respectively (Fig.~\ref{fig:morphology_robustness}). These results support transfer beyond the original link dimensions through functional correspondence, without retraining.

\subsection{Diversity Analysis}
\textbf{Diversity.} \method{} achieves a diversity of $0.293$~rad, below the paper-reported $0.401$~rad for \trograsp{}, $0.450$~rad for \drograsp{}, and $0.512$~rad for CEDex (Tab.~\ref{tab:diversity}). The generated spatial arrangements of the functional parts already show limited variation before inverse kinematics, suggesting that the limited diversity originates at least partly in grasp generation.

\begin{table}[t]
\centering
\footnotesize
\setlength{\tabcolsep}{3.5pt}
\renewcommand{\arraystretch}{1.2}
\begin{tabular*}{\columnwidth}{@{\extracolsep{\fill}}lccccc@{}}
\toprule
\textbf{Hand} & Ball & Block & Bread & Cracker & Drill \\
\midrule
ApexHand & 10/10 & 10/10 & 9/10 & 9/10 & 7/10  \\
Revo2    & 9/10  & 8/10  & 9/10 & 6/10 & 5/10  \\
\midrule
& Vase & Milk & Mustard & Pepper & Soup  \\
\midrule
ApexHand & 5/10 & 8/10 & 8/10 & 10/10 & 8/10  \\
Revo2    & 4/10 & 8/10 & 6/10 & 8/10 & 5/10  \\
\bottomrule
\end{tabular*}
\caption{Per-object real-world grasp success on ApexHand and Revo2, both unseen during training. Each entry reports successful grasps out of ten trials.}
\label{tab:real_world}
\vspace*{-0.15in}
\end{table}

\begin{figure}[t]
  \centering
  \includegraphics[width=\linewidth,clip=true]{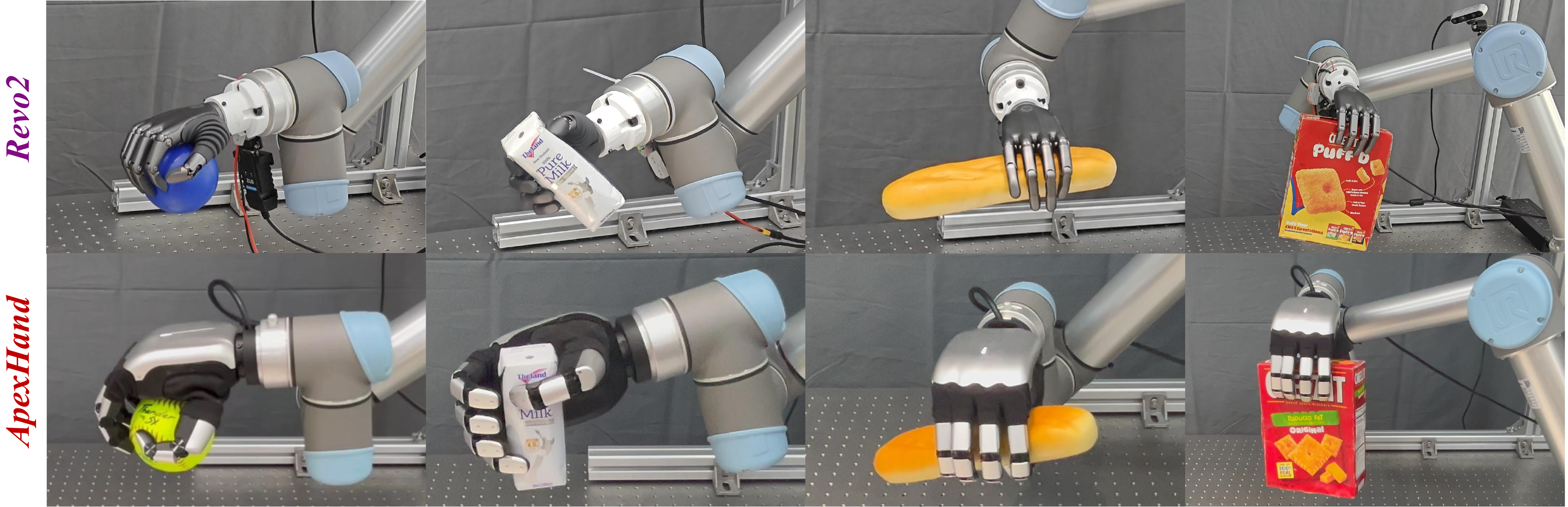}
  \caption{Representative successful grasps on Revo2 (top) and ApexHand (bottom). The same model is deployed on both hands without additional training or fine-tuning.}
  \label{fig:realexe2}
  \vspace*{-0.2in}
\end{figure}

\subsection{Real-World Experiments}

\textbf{Hardware.} Our real-world platforms use UR5 arms equipped with ApexHand or Revo2 and an Intel RealSense D435i (Fig.~\ref{fig:real_setup}). We use the same model evaluated in Tab.~\ref{tab:overall} to generate grasps for both hands without additional training or fine-tuning. Both hands are unseen during training.

\textbf{Execution Protocol.} Each trial resets the arm, opens the hand, and captures fresh RGB-D data. We segment the target object in the RGB image using SAM2~\cite{ravi2024sam2} and use the corresponding depth data to obtain its point cloud for grasp generation. Each hand is evaluated across ten random placements for each of the 10 objects, totaling 100 trials per hand. Success requires lifting the object clear of the table and maintaining a stable grasp.

\textbf{Real-World Results.} ApexHand and Revo2 achieve 84/100 and 68/100 successful grasps, respectively, yielding $76.00\%$ overall (Tab.~\ref{tab:real_world}). Per-object results range from 5/10 to 10/10 for ApexHand and 4/10 to 9/10 for Revo2, with representative successful grasps in Fig.~\ref{fig:realexe2}. Real-world experiments further demonstrate that the same model transfers grasp knowledge to two unseen hands without additional training or fine-tuning.

\section{Conclusion}
We present \method{} for cross-embodiment dexterous grasp generation through functional correspondences across heterogeneous hands. Shared functional parts and canonical local frames provide a consistent representation for inter-part and hand-object interactions, allowing the model to learn transferable grasp knowledge. Simulation experiments demonstrate transfer to unseen hands without target-hand grasp data, fine-tuning, or learned retargeting. Real-world experiments further demonstrate that the same model transfers grasp knowledge to two unseen hands without additional training or fine-tuning. Grasp diversity remains below the paper-reported baseline values, motivating future work on task-conditioned grasp generation guided by human demonstrations.



\bibliographystyle{IEEEtran}
\bibliography{tex/root}

@inproceedings{kim2025pyroki,
  author    = {Chung Min Kim and Brent Yi and Hongsuk Choi and Yi Ma and Ken Goldberg and Angjoo Kanazawa},
  title     = {{PyRoki}: A Modular Toolkit for Robot Kinematic Optimization},
  booktitle = {2025 IEEE/RSJ International Conference on Intelligent Robots and Systems (IROS)},
  year      = {2025},
  url       = {https://arxiv.org/abs/2505.03728}
}

@article{shao2020unigrasp,
  author  = {Lin Shao and Fabio Ferreira and Mikael Jorda and Varun Nambiar and Jianlan Luo and Eugen Solowjow and Juan Aparicio Ojea and Oussama Khatib and Jeannette Bohg},
  title   = {UniGrasp: Learning a Unified Model to Grasp With Multifingered Robotic Hands},
  journal = {IEEE Robotics and Automation Letters},
  volume  = {5},
  number  = {2},
  pages   = {2286--2293},
  year    = {2020}
}

@inproceedings{attarian2023geomatch,
  author    = {Maria Attarian and Muhammad Adil Asif and Jingzhou Liu and Ruthrash Hari and Animesh Garg and Igor Gilitschenski and Jonathan Tompson},
  title     = {Geometry Matching for Multi-Embodiment Grasping},
  booktitle = {Proceedings of the Conference on Robot Learning (CoRL)},
  series    = {Proceedings of Machine Learning Research},
  volume    = {229},
  pages     = {1242--1256},
  year      = {2023}
}

@inproceedings{li2023gendexgrasp,
  author    = {Puhao Li and Tengyu Liu and Yuyang Li and Yiran Geng and Yixin Zhu and Yaodong Yang and Siyuan Huang},
  title     = {GenDexGrasp: Generalizable Dexterous Grasping},
  booktitle = {Proceedings of the IEEE International Conference on Robotics and Automation (ICRA)},
  pages     = {8068--8074},
  year      = {2023}
}

@article{liu2022dfc,
  author  = {Tengyu Liu and Zeyu Liu and Ziyuan Jiao and Yixin Zhu and Song-Chun Zhu},
  title   = {Synthesizing Diverse and Physically Stable Grasps With Arbitrary Hand Structures Using Differentiable Force Closure Estimator},
  journal = {IEEE Robotics and Automation Letters},
  volume  = {7},
  number  = {1},
  pages   = {470--477},
  year    = {2022},
  doi     = {10.1109/LRA.2021.3129138}
}

@inproceedings{wei2025dro,
  author    = {Zhenyu Wei and Zhixuan Xu and Jingxiang Guo and Yiwen Hou and Chongkai Gao and Zhehao Cai and Jiayu Luo and Lin Shao},
  title     = {$\mathcal{D}(\mathcal{R},\mathcal{O})$ Grasp: A Unified Representation of Robot and Object Interaction for Cross-Embodiment Dexterous Grasping},
  booktitle = {Proceedings of the IEEE International Conference on Robotics and Automation (ICRA)},
  year      = {2025}
}

@article{fei2025tro,
  author  = {Xin Fei and Zhixuan Xu and Huaicong Fang and Tianrui Zhang and Lin Shao},
  title   = {$\mathcal{T}(\mathcal{R},\mathcal{O})$ Grasp: Efficient Graph Diffusion of Robot-Object Spatial Transformation for Cross-Embodiment Dexterous Grasping},
  journal = {arXiv preprint arXiv:2510.12724},
  year    = {2025}
}

@inproceedings{yuan2025crossdex,
  author    = {Haoqi Yuan and Bohan Zhou and Yuhui Fu and Zongqing Lu},
  title     = {Cross-Embodiment Dexterous Grasping with Reinforcement Learning},
  booktitle = {International Conference on Learning Representations (ICLR)},
  year      = {2025}
}

@article{patel2024getzero,
  author  = {Austin Patel and Shuran Song},
  title   = {GET-Zero: Graph Embodiment Transformer for Zero-shot Embodiment Generalization},
  journal = {arXiv preprint arXiv:2407.15002},
  year    = {2024}
}

@article{wu2026dexgraspzero,
  author  = {Yuliang Wu and Yanhan Lin and WengKit Lao and Yuhao Lin and Yi-Lin Wei and Wei-Shi Zheng and Ancong Wu},
  title   = {DexGrasp-Zero: A Morphology-Aligned Policy for Zero-Shot Cross-Embodiment Dexterous Grasping},
  journal = {arXiv preprint arXiv:2603.16806},
  year    = {2026}
}

@article{wu2026unimorph,
  author  = {Zhiyuan Wu and Xiangyu Zhang and Zhuo Chen and Jiankang Deng and Rolandos Alexandros Potamias and Shan Luo},
  title   = {UniMorphGrasp: Diffusion Model with Morphology-Awareness for Cross-Embodiment Dexterous Grasp Generation},
  journal = {arXiv preprint arXiv:2602.00915},
  year    = {2026}
}

@article{wu2025cedex,
  author  = {Zhiyuan Wu and Rolandos Alexandros Potamias and Xuyang Zhang and Zhongqun Zhang and Jiankang Deng and Shan Luo},
  title   = {CEDex: Cross-Embodiment Dexterous Grasp Generation at Scale from Human-like Contact Representations},
  journal = {arXiv preprint arXiv:2509.24661},
  year    = {2025}
}

@article{zhang2025machagrasp,
  author  = {Heng Zhang and Kevin Yuchen Ma and Mike Zheng Shou and Weisi Lin and Yan Wu},
  title   = {MachaGrasp: Morphology-Aware Cross-Embodiment Dexterous Hand Articulation Generation for Grasping},
  journal = {arXiv preprint arXiv:2510.06068},
  year    = {2025}
}

@article{lee2026graspgraphnet,
  author  = {Yeonseo Lee and Taeyeop Lee and Hyosup Shin and Guebin Hwang and Sungho Jo},
  title   = {GraspGraphNet: Graph-Structured Multi-Embodiment Dexterous Grasp Generation},
  journal = {arXiv preprint arXiv:2607.11031},
  year    = {2026}
}

@article{zhang2026mango,
  author  = {Heng Zhang and Kevin Yuchen Ma and Mike Zheng Shou and Weisi Lin and Yan Wu},
  title   = {MANGO-Grasp: Mahalanobis Fields over Geometry-Oriented 3D Gaussians for Cross-Embodiment Dexterous Grasping},
  journal = {arXiv preprint arXiv:2608.02014},
  year    = {2026}
}

@inproceedings{wang2023dexgraspnet,
  author    = {Ruicheng Wang and Jialiang Zhang and Jiayi Chen and Yinzhen Xu and Puhao Li and Tengyu Liu and He Wang},
  title     = {DexGraspNet: A Large-Scale Robotic Dexterous Grasp Dataset for General Objects Based on Simulation},
  booktitle = {Proceedings of the IEEE International Conference on Robotics and Automation (ICRA)},
  year      = {2023}
}

@inproceedings{xu2023unidexgrasp,
  author    = {Yinzhen Xu and Weikang Wan and Jialiang Zhang and Haoran Liu and Zikang Shan and Hao Shen and Ruicheng Wang and Haoran Geng and Yijia Weng and Jiayi Chen and Tengyu Liu and Li Yi and He Wang},
  title     = {UniDexGrasp: Universal Robotic Dexterous Grasping via Learning Diverse Proposal Generation and Goal-Conditioned Policy},
  booktitle = {Proceedings of the IEEE/CVF Conference on Computer Vision and Pattern Recognition (CVPR)},
  pages     = {4737--4746},
  year      = {2023}
}

@inproceedings{xu2024dgtr,
  author    = {Guo-Hao Xu and Yi-Lin Wei and Dian Zheng and Xiao-Ming Wu and Wei-Shi Zheng},
  title     = {Dexterous Grasp Transformer},
  booktitle = {Proceedings of the IEEE/CVF Conference on Computer Vision and Pattern Recognition (CVPR)},
  pages     = {17933--17942},
  year      = {2024}
}

@inproceedings{zhong2025dexgraspanything,
  author    = {Yiming Zhong and Qi Jiang and Jingyi Yu and Yuexin Ma},
  title     = {DexGrasp Anything: Towards Universal Robotic Dexterous Grasping with Physics Awareness},
  booktitle = {Proceedings of the IEEE/CVF Conference on Computer Vision and Pattern Recognition (CVPR)},
  year      = {2025}
}

@inproceedings{turpin2022graspd,
  author    = {Dylan Turpin and Liquan Wang and Eric Heiden and Yun-Chun Chen and Miles Macklin and Stavros Tsogkas and Sven Dickinson and Animesh Garg},
  title     = {Grasp'D: Differentiable Contact-Rich Grasp Synthesis for Multi-Fingered Hands},
  booktitle = {Proceedings of the European Conference on Computer Vision (ECCV)},
  year      = {2022}
}

@inproceedings{brahmbhatt2019contactdb,
  author    = {Samarth Brahmbhatt and Cusuh Ham and Charles C. Kemp and James Hays},
  title     = {ContactDB: Analyzing and Predicting Grasp Contact via Thermal Imaging},
  booktitle = {Proceedings of the IEEE/CVF Conference on Computer Vision and Pattern Recognition (CVPR)},
  pages     = {8709--8719},
  year      = {2019}
}

@inproceedings{calli2015ycb,
  author    = {Berk Calli and Arjun Singh and Aaron Walsman and Siddhartha Srinivasa and Pieter Abbeel and Aaron M. Dollar},
  title     = {The YCB Object and Model Set: Towards Common Benchmarks for Manipulation Research},
  booktitle = {Proceedings of the International Conference on Advanced Robotics (ICAR)},
  pages     = {510--517},
  year      = {2015}
}

@inproceedings{qi2017pointnet,
  author    = {Charles R. Qi and Hao Su and Kaichun Mo and Leonidas J. Guibas},
  title     = {PointNet: Deep Learning on Point Sets for 3D Classification and Segmentation},
  booktitle = {Proceedings of the IEEE Conference on Computer Vision and Pattern Recognition (CVPR)},
  pages     = {652--660},
  year      = {2017}
}

@inproceedings{ho2020ddpm,
  author    = {Jonathan Ho and Ajay Jain and Pieter Abbeel},
  title     = {Denoising Diffusion Probabilistic Models},
  booktitle = {Advances in Neural Information Processing Systems (NeurIPS)},
  volume    = {33},
  pages     = {6840--6851},
  year      = {2020}
}

@inproceedings{song2021ddim,
  author    = {Jiaming Song and Chenlin Meng and Stefano Ermon},
  title     = {Denoising Diffusion Implicit Models},
  booktitle = {International Conference on Learning Representations (ICLR)},
  year      = {2021}
}

@inproceedings{makoviychuk2021isaacgym,
  author    = {Viktor Makoviychuk and Lukasz Wawrzyniak and Yunrong Guo and Michelle Lu and Kier Storey and Miles Macklin and David Hoeller and Nikita Rudin and Arthur Allshire and Ankur Handa and Gavriel State},
  title     = {Isaac Gym: High Performance GPU-Based Physics Simulation for Robot Learning},
  booktitle = {Advances in Neural Information Processing Systems Datasets and Benchmarks Track},
  year      = {2021}
}

@inproceedings{ravi2024sam2,
  author    = {Ravi, Nikhila and Gabeur, Valentin and Hu, Yuan-Ting and Hu, Ronghang and Ryali, Chaitanya and Ma, Tengyu and Khedr, Haitham and R\"{a}dle, Roman and Rolland, Chloe and Gustafson, Laura and Mintun, Eric and Pan, Junting and Alwala, Kalyan Vasudev and Carion, Nicolas and Wu, Chao-Yuan and Girshick, Ross and Dollar, Piotr and Feichtenhofer, Christoph},
  title     = {{SAM 2}: Segment Anything in Images and Videos},
  booktitle = {International Conference on Learning Representations},
  pages     = {28085--28128},
  year      = {2025},
  url       = {https://proceedings.iclr.cc/paper_files/paper/2025/file/45c1f6a8cbf2da59ebf2c802b4f742cd-Paper-Conference.pdf}
}

\end{document}